\documentclass[10pt, a4paper]{article}
\usepackage{lrec2022} 
\usepackage{multibib}
\newcites{languageresource}{Language Resources}
\usepackage{graphicx}
\usepackage{tabularx}
\usepackage{booktabs}
\usepackage{soul}
\usepackage{titlesec}
\titleformat{\section}{\normalfont\large\bfseries\center}{\thesection.}{1em}{}
\titleformat{\subsection}{\normalfont\SmallTitleFont\bfseries\raggedright}{\thesubsection.}{1em}{}
\titleformat{\subsubsection}{\normalfont\normalsize\bfseries\raggedright}{\thesubsubsection.}{1em}{}
\renewcommand\thesection{\arabic{section}}
\renewcommand\thesubsection{\thesection.\arabic{subsection}}
\renewcommand\thesubsubsection{\thesubsection.\arabic{subsubsection}}

\usepackage{epstopdf}
\usepackage[utf8]{inputenc}

\usepackage{hyperref}
\usepackage{xstring}

\usepackage{color}

\title{EENLP: Cross-lingual Eastern European NLP Index}
\name{
    \\\textbf{Alexey Tikhonov\textsuperscript{*}}\thanks{* The corresponding author is Alexey Tikhonov}, \textbf{Alex Malkhasov}, \textbf{Andrey Manoshin}, \textbf{George Dima}, \\\textbf{Réka Cserháti}, \textbf{Md.Sadek Hossain Asif}, \textbf{Matt Sárdi}\\
}
\address{
         Independent researcher, Germany; Financial University of Russia; MEPhI, Russia; \\
         University Politehnica of Bucharest, Romania; University of Szeged, Hungary; \\
         Notre Dame College, Dhaka, Bangladesh; Mozaik Education, Hungary
  \\\\
         \{altsoph, realex1902, sqidde, andreigeorgedima, cserhatir, asifsadek509, sardi.matt\}@gmail.com\\
}

\abstract{
Motivated by the sparsity of NLP resources for Eastern European languages, we present a broad index of existing Eastern European language resources (90$+$ datasets and 45$+$ models) published as a github repository open for updates from the community. Furthermore, to support the evaluation of commonsense reasoning tasks, we provide hand-crafted cross-lingual datasets for five different semantic tasks (namely news categorization, paraphrase detection, Natural Language Inference (NLI) task, tweet sentiment detection, and news sentiment detection) for some of the Eastern European languages. We perform several experiments with the existing multilingual models on these datasets to define the performance baselines and compare them to the existing results for other languages.
 \\ \newline  
}

\begin{document}

\maketitleabstract

\section{Introduction}

Recent multilingual Transformer-based language models  -- such as mBERT \cite{devlin2018bert}, XLM-RoBERTa \cite{Sanh2019DistilBERTAD}, multilingual DistilBERT \cite{conneau2020unsupervised}, etc. -- show impressive results on different text analysis tasks and their cross-lingual reasoning capabilities are still actively studied \cite{lauscher-etal-2020-zero}. For example, the mBERT model trained on 104 languages has shown high cross-lingual performance; however, such evaluations mostly focused on cross-lingual transfer within high-resource languages \cite{wu-dredze-2020-languages}. 
\\\\
Commonsense reasoning is one of the key problems in natural language processing, but the relative scarcity of labeled data holds back the progress for languages other than English: there are widely spoken languages that still did not receive the focus of the research community \cite{joshi-etal-2020-state}. Cross-lingual transfer learning could be beneficial for such languages in solving both theoretical and practical tasks. One can speculate that such high-level reasoning tasks could be less affected by the language syntax's specifics and low-level properties, so it can be effective to use a cross-lingual approach here. However, it was shown that the success of cross-lingual transfer learning depends on different factors such as the amount of shared vocabulary, explicit alignment of representations across languages, size of pretraining corpora, etc \cite{doddapaneni2021primer}. To get a better understanding of the importance of these factors, researchers need to leverage diverse and detailed datasets.
\\\\
There are a bunch of cross-lingual datasets already, such as XGLUE \cite{XGLUE}, XCOPA \cite{XCOPA}, XL-WiC \cite{XL-WiC}, XWINO \cite{XWINE}, Wino-X \cite{emelin-sennrich-2021-wino}, XNLI \cite{XNLI}, XL-WSD \cite{XL-WSD}, XTREME-R \cite{ruder-etal-2021-xtreme}, BSNLP \cite{BSNLP}, MOROCO \cite{MOROCO}, etc., but most of them cover high-resource languages.
\\\\
In this paper, we concentrate on Eastern European languages. These languages are numerous and heterogeneous; they include languages from at least two language families (Indo-European and Uralic), and the former family is represented by very diverse branches. Although there are several dedicated multi-task benchmarks for a few Eastern European languages -- e.g., KLEJ \cite{KLEJ} for Polish, LiRo \cite{LiRo} for Romanian, RussianSuperGLUE \cite{RussianSuperGLUE} for Russian, or the translation of SuperGLUE for Slovene, they usually concentrate on one or two languages. They also use different tasks and various data formats; hence they can not be used for cross-lingual benchmarking without careful manual pre-processing. 
\\\\
The main contributions of this paper are:
\begin{itemize}
    \item{To build a comprehensive picture of the current NLP state for the Eastern European languages, we present the wide index of existing Eastern European languages resources (90+ datasets and 45+ models) published as a github repository\footnote{\scriptsize{https://github.com/altsoph/EENLP}} open for updates from the community;}
    \item{Next, we provide hand-crafted cross-lingual datasets for five different semantic tasks (namely news categorization, paraphrase detection, Natural Language Inference (NLI) task, tweet sentiment detection, and news sentiment detection), compiled by processing data from various sources into the same format, opening room for evaluation scenarios such as zero-shot cross-lingual transfer. Since the source datasets are licensed under various licenses, we publish automatic scripts for our datasets compilation on the same github repository;}
    \item{Finally, we perform several experiments with the existing multilingual models on our datasets to define the performance baselines and compared them to the existing similar results for other languages.}
\end{itemize}
We made all our code and data publicly available. We also published the detailed results of our experiments at our Weights and Biases project\footnote{\scriptsize{https://wandb.ai/eenlp}} \cite{wandb}. 
\begin{figure*}[h!]\centering
     \includegraphics[scale=0.36]{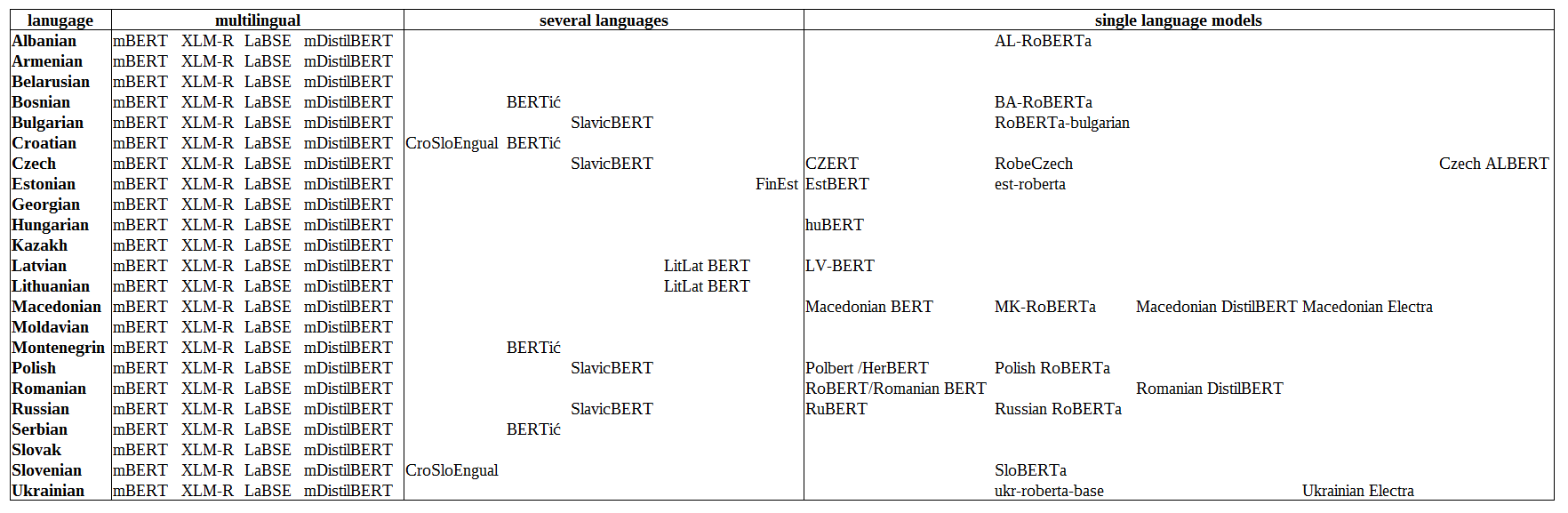}
  \caption{MLM Transformer-based models with Eastern European languages coverage.}
  \label{fig:mlm}
\end{figure*}
\section{Dataset index}
Aiming to build the index of Eastern European language NLP resources useful to the community, we have created a long list of available datasets we were able to find. We focused on supervised datasets (with data labels) and preferred semantic tasks to syntactic ones. Lastly, since some languages are already well covered with benchmarks (for example, Polish, Romanian, Russian), we tried to focus on less represented languages. Indeed, we do not claim this list is exhaustive; however, we provide an easy way to add any missing resources via the creation of a github issue. We encourage the community to help with updating this index further.
\\\\
For this moment, we've already collected more than 90 datasets for 20 different languages: Albanian, Armenian, Belarusian, Bosnian, Bulgarian, Croatian, Czech, Estonian, Hungarian, Latvian, Lithuanian, Macedonian, Moldavian, Polish, Romanian, Russian, Serbian, Slovak, Slovenian, and Ukrainian. 
\\\\
These datasets cover various tasks, including text category prediction, coreference resolution, fake news detection, lemmatization, morphosyntactic tagging, NER, NLI, offensive comments detection, paraphrase detection, POS tagging, question answering, sentiment analysis, word sense disambiguation, and many more. The complete list of the discovered datasets is published on our github repository\footnote{\scriptsize{https://github.com/altsoph/EENLP/blob/main/docs/datasets.md}}.

\section{Models index}
In a similar way, we tried to enumerate available models useful for Eastern European languages. We were aiming for the Transformer-based MLM models (which are state of the art for text classification nowadays), but we also found some models of other types, including Causal Language Models, Adapters, static embeddings, etc; we listed them as well.
\\\\
We started with the modern well-known multilingual models, namely mBERT \cite{devlin2018bert}, mDistilBERT \cite{Sanh2019DistilBERTAD}, XLM-Roberta \cite{conneau2020unsupervised}, and LaBSE \cite{LaBSE}. Next, there are several cross-lingual models, such as CroSloEngual \cite{CroSlo}, BERTić \cite{BERTIC}, SlavicBERT \cite{SlavicBERT}, or LitLat BERT. Finally, we found more than 20 language-specific BERT-like models; most of them are listed in Figure \ref{fig:mlm}.
\\\\
Besides MLM models, we found 6 GPT-like models and several dozens of models from the pre-Transformer era: ULMFiT, ELMo, and static word embeddings. All our findings are available on our github repository\footnote{\scriptsize{https://github.com/altsoph/EENLP/blob/main/docs/models.md}} as well.
\section{Benchmark tasks}
Considering the coverage and sparsity of source datasets, we decided to proceed with five commonsense reasoning tasks with different Eastern European languages subsets. We checked these tasks for the languages coverage and finally decided to focus on this short-list: 
\vspace{-4pt}
\begin{itemize}
\setlength{\itemsep}{0pt}
    \item news categorisation,
    \item paraphrase detection, 
    \item news sentiment detection, 
    \item tweet sentiment detection,
    \item NLI.
\end{itemize}

For each of these tasks, we manually crafted a dedicated dataset with maximized Eastern European languages coverage. Whenever it was possible we also added English as a language most common for contemporary pre-trained models. The properties of these datasets are given in the following sections. Since the source datasets are licensed under various licenses, we published automatic scripts for our datasets compilation on the same github repository\footnote{\scriptsize{https://github.com/altsoph/EENLP/tree/main/build\_benchmarks}}.
\\\\
For each of these tasks, we evaluated three modern transformer-based pre-trained models: a multilanguage version of BERT, a model XLM-Roberta that is two times larger than BERT, and a distilled version of multilanguage BERT (to check how the distillation affects the cross-lingual transfer quality). Whenever the English language was also available for the task we added a base English BERT model to our evaluation. We fine-tuned them using AdamW optimizer with the learning rate 1e-5, epsilon 1e-8, and the linear schedule with warmup. 
We reported all the metrics for the epoch with the best source language validation result and published our code for evaluation\footnote{\scriptsize{https://github.com/altsoph/EENLP/tree/main/eval\_benchmarks}}.
\\\\
We will now describe every task in detail.

\subsection{News categorisation task}

In this task, the model should detect the correct category of the news text. This is a basic semantic task; however, it's not always obvious how to separate, for example, categories like "lifestyle" and "entertainment." Our dataset consists of 8 major categories and covers 7 Eastern European languages + English (however, the distribution of categories differs from one language to another, consider Table \ref{tab:task_a_data}).
\\\\
The sources for this dataset are:
\begin{itemize}
\setlength{\itemsep}{0pt}
    \item English: \cite{XGLUE},
    \item Armenian: \cite{nc_armenian},
    \item Estonian: \cite{nc_ee},
    \item Latvian: \cite{nc_lv},
    \item Moldavian: \cite{MOROCO},
    \item Romanian: \cite{MOROCO},
    \item Russian: \cite{XGLUE},
    \item Slovak: \cite{hladek-etal-2014-slovak}.
\end{itemize}

\begin{table*}[]
\centering
\begin{tabular}{lrrrrrrrr}
 & \textbf{en} & \textbf{hy} & \textbf{ee} & \textbf{lv} & \textbf{mo} & \textbf{ro} & \textbf{ru} & \textbf{sk} \\
\hline
entertainment & 3498 & 996 & 275 & 227 & 1234 & 1058 & 1498 & 22 \\
finance & 13405 & 2066 & 13 & 34 & 3429 & 5105 & 2514 & 1022 \\
health  & 5751 &  & 121 & 119 &  &  & 311 & 559 \\
lifestyle  & 6295 & 1630 & 107 & 37 &  &  & 671 & 481 \\
news/accidents  & 31490 & 1454 & 33526 & 3722 &  &  & 12125 & 1655 \\
sports  & 38598 & 2797 &  & 149 & 3443 & 2583 & 1731 & 1042 \\
tech/auto  & 3926 &  & 144 & 40 & 1266 & 3392 & 387 &  \\
travel  & 3076 &  & 49 & 55 &  &  & 53 &  \\
\end{tabular}
\caption{News categorisation dataset properties. For each language / category pair we reported the number of the items (short news texts) in our dataset.}
\label{tab:task_a_data}
\end{table*}

We use this task to measure the models' cross-lingual transfer quality while transferring from English to various Eastern European languages. (One can also use Latvian or Russian as a source language since both of them cover all of the categories). Following \cite{XGLUE}, we use multi-class classification accuracy as the key metric for this task. Table \ref{tab:task_a_res} sums up the results of four models on a multilanguage dataset. Further details across languages and categories are available on W\&B\footnote{https://wandb.ai/eenlp/newscat\_full}.
\\\\
\begin{table}[]
\begin{tabular}{lllll}
                                     & mDBERT & mBERT & XLMR & BERT \\
\hline
en                       & 91.11\%  & \textbf{92.50}\% & 92.36\% &  92.65\% \\
ee                       & 74.83\%  & \textbf{81.58}\% & 72.55\% & 27.33\% \\
hy                       & 55.57\%  & 66.10\% & \textbf{67.64\%} & 21.41\% \\
lv                       & 63.79\%  & 86.60\% & \textbf{87.68\%} & 6.31\% \\
mo                       & 65.79\%  & \textbf{66.11}\% & 62.36\% & 16.12\% \\
ro                       & 46.18\%  & \textbf{47.54}\% & 44.18\% & 7.46\% \\
ru                       & 77.49\%  & \textbf{77.34}\% & 77.10\% & 43.31\% \\
sk                       & 59.88\%  & \textbf{63.25}\% & 60.32\% & 35.34\% \\
\end{tabular}
\caption{News categorisation task. For each language / model pair we reported the accuracy value on the same epoch where the validation accuracy was the best. We also evaluated a standard English BERT here to compare.}
\label{tab:task_a_res}
\end{table}
The main observations based on the results of this task are:
 \begin{itemize}
    \item multilingual pre-training is the key factor for successful cross-lingual transfer learning, English BERT model shows no ability to transfer its knowledge from English to other languages;
    \item mBERT and XLM-R models show a similar level of quality as in \cite{XGLUE} for the similar NC task (however, note, it's unfair to compare them directly since the list of categories is not the same);
    \item distillation significantly affects the ability to cross-lingual transfer learning; this effect is significant for almost all of the tested languages;
    \item the results vary across the languages a lot, this is a direction of possible deeper analysis (note that, for example, Romanian and Moldavian are close languages; however, all the models are performing significantly better on Moldavian);
    \item mBERT generally performs slightly better than XLM-RoBERTa, this could be an effect of relatively small datasets.
\end{itemize}    

\subsection{Paraphrase detection task}
\begin{table}[h!]
\centering
\begin{tabular}{lrr}
 Language & Items & Classes   \\ \hline
 \begin{tabular}[c]{@{}l@{}}English\end{tabular}   & 373 263 & 2 \\
 \begin{tabular}[c]{@{}l@{}}Armenian\end{tabular}   & 4 233 & 2 \\
 \begin{tabular}[c]{@{}l@{}}Polish\end{tabular}   & 8 000 & 2 \\
 \begin{tabular}[c]{@{}l@{}}Romanian\end{tabular}   & 5 749 & 2 \\
 \begin{tabular}[c]{@{}l@{}}Serbian\end{tabular}   & 835 & 2 \\
\end{tabular}
  \caption{Paraphrase detection dataset properties. Each item is a pair of sentences.}
  \label{tab:task_pd_stat}
\end{table}

In the paraphrase detection task, the model should decide whether a pair of sentences have the same meaning. It is another standard task for contemporary NLU benchmarks; however, it is significantly hard and can be problematic even for the English language \cite{semsim}. Our dataset consists of four Eastern European languages + English. The sources for this dataset are: 
\begin{itemize}
\setlength{\itemsep}{0pt}
    \item English: \cite{GLUE},
    \item Armenian: \cite{arpa},
    \item Polish: \cite{KLEJ},
    \item Romanian: RO-STS dataset\footnote{https://github.com/dumitrescustefan/RO-STS},
    \item Serbian: \cite{pd_sr}.
\end{itemize}

We also tried to leverage the TaPaCo dataset \cite{TaPaCo} to build a derivative dataset with positive and negative examples of paraphrases. We treated pairs of sentences from the same cluster as positive paraphrase examples. Then, we used several strategies for sampling hard-negative samples (including sampling from similar sentences by Levenshtein distance and LabSE \cite{labse_ref} embedding distances). To assess the quality of such dataset augmentation, we performed a two-folded cross-check: considering languages where we have both TaPaCo and non-TaPaCo datasets, we checked the cross-datasets transfer learning in both directions. The results of non-TaPaCo-$>$TaPaCo transfer were much better than for the opposite direction (for example, for Polish, it was 68\% versus 51\%), which implies that datasets represent different tasks with non-equal complexity. Thus, we decided to reject a TaPaCo-based augmentation until further investigations. However, we consider it an interesting direction for future work.
\\\\
We used the same setup as in the previous task -- cross-lingual zero-shot learning from English to other languages. Following \cite{PAWSX}, we use classification accuracy as the key metric for this task. We reported values in table \ref{tab:task_pd_res} from the same epoch where the validation accuracy was the best. More details are available online\footnote{https://wandb.ai/eenlp/paraphrase\_detection/artifacts/\\paraphrase\_detection-dataset/paraphrase\_detection-dataset/ d144e76e6129b022dc5d}.
\\\\
\begin{table}[]
\begin{tabular}{lllll}
                                     & mDBERT & mBERT & XLMR & BERT \\
\hline
en                       & 81.84\%  & 82.20\% & \textbf{83.80}\% & 82.20\% \\

hy                       & 88.68\%  & \textbf{92.79}\% & 92.70\% & 75.36\% \\
pl                       & 87.86\%  & \textbf{87.29}\% & 82.14\% & 84.33\% \\
ro                       & 79.82\%  & 82.36\% & \textbf{86.88}\% & 80.03\% \\
sr                       & 72.10\%  & \textbf{77.01}\% & 71.74\% & 55.09\% \\

\end{tabular}
\caption{Paraphrase detection task. For each language / model pair we reported the accuracy value on the same epoch where the validation accuracy was the best. We also evaluated a standard English BERT here to compare.}
\label{tab:task_pd_res}
\end{table}

The main observations based on these results are:
 \begin{itemize}
    \item again, multilingual pre-trained models show consistently better performance than English BERT, which highlights the importance of the multilingual pre-training phase;
    \item mBERT results are competitive with the results reported in \cite{PAWSX};
    \item once more, distillation affects the ability to cross-lingual transfer learning.
\end{itemize}

\subsection{News sentiment detection task}
\begin{table}[h!]
\centering
\begin{tabular}{lrr}
 Language & Items & Classes   \\ \hline
 \begin{tabular}[c]{@{}l@{}}Croatian\end{tabular}   & 1 500 & 3 \\
 \begin{tabular}[c]{@{}l@{}}Lithuanian\end{tabular}   & 10 000 & 3 \\
 \begin{tabular}[c]{@{}l@{}}Russian\end{tabular}   & 8 200 & 3 \\
 \begin{tabular}[c]{@{}l@{}}Slovenian\end{tabular}   & 10 000 & 3 \\
\end{tabular}
  \caption{News sentiment detection dataset properties. Each item is a short news text.}
  \label{task_san_data}
\end{table}

In this task, we made an evaluation of sentiment detection (3 classes: positive, neutral, negative) for relatively long texts (news articles). We compiled a dataset, that covers 4 Eastern European languages (Croatian, Lithuanian, Russian, and Slovenian). Table \ref{task_san_data} describes volumes of data per language. The sources for this dataset are: 
\begin{itemize}
\setlength{\itemsep}{0pt}
    \item Croatian: \cite{nsczh},
    \item Lithuanian: TN2gramos \footnote{https://www.kaggle.com/rokastrimaitis/lithuanian-
    financial-news-dataset-and-bigrams},
    \item Russian: Sentiment Analysis in Russian\footnote{https://www.kaggle.com/c/sentiment-analysis-in-russian},
    \item Slovenian: \cite{Slovenian_nsd}.
\end{itemize}

We used classification accuracy as our main metric and evaluated all possible directions of transfer learning for three multilingual models (mBERT, mDBERT, and XLM-R), as in \cite{nsczh}. Check table \ref{task_san_res} for the results, more details are available on W\&B\footnote{https://wandb.ai/eenlp/news\_sentiment}.
\\\\
\begin{table}[]
\begin{tabular}{llllll}
& & \textbf{hr}& \textbf{lt}& \textbf{ru}& \textbf{sl}\\
& mDBERT& 67\%& 20\%& 48\%& 53\%\\
\textbf{hr}& mBERT& 66\%& 23\%& 53\%& 57\%\\
& XLMR& \textbf{70}\%& \textbf{24}\%& \textbf{50}\%& \textbf{60}\%\\
& mDBERT& 45\%& 70\%& 50\%& 51\%\\
\textbf{lt}& mBERT& 49\%& 71\%& 57\% & 49\%\\
& XLMR& \textbf{51}\%& \textbf{78}\%& \textbf{58}\%& \textbf{63}\%\\
& mDBERT& \textbf{63}\%& 39\%& 68\%& 52\%\\
\textbf{ru}& mBERT& 57\%& \textbf{39}\%& \textbf{71}\%& 55\%\\
& XLMR& 56\%& 65\%& 70\%& \textbf{59}\%\\
& mDBERT& \textbf{55}\%& 41\%& 57\%& 64\%\\
\textbf{sl}& mBERT& 54\%& 52\%& 58\%& 67\%\\
& XLMR& 51\%& \textbf{59}\%& \textbf{60}\%& \textbf{69}\%\\
\end{tabular}
\caption{News sentiment detection task. Validation accuracy.}
\label{task_san_res}
\end{table}
This task results support several observations:
 \begin{itemize}
    \item as in previous experiments, the distillation decreases the ability to cross-lingual transfer learning;
    \item the cross-lingual transfer quality seems to be correlated with the degree of language affinity;
    \item XLM-RoBERTa generally dominates over the rest of the models.
\end{itemize}

\subsection{Twitter sentiment detection task}

\begin{table}[h!]
\centering
\begin{tabular}{lrr}
 Language & Items & Classes   \\ \hline
 \begin{tabular}[c]{@{}l@{}}Czech\end{tabular}   & 10 000 & 3 \\
 \begin{tabular}[c]{@{}l@{}}Latvian\end{tabular}   & 1 160 & 3 \\
 \begin{tabular}[c]{@{}l@{}}Russian\end{tabular}   & 230 000 & 2 \\
 \begin{tabular}[c]{@{}l@{}}Slovak\end{tabular}   & 1 600 & 3 \\
 
\end{tabular}
  \caption{Twitter sentiment detection dataset properties. Each item is a tweet text.}
  \label{task_sat_data}
\end{table}

In addition to the previous task, we considered the task of short texts (tweets) sentiment detection (same 3 classes: positive, neutral, negative). The dataset for this task also covers 4 Eastern European languages (Czech, Latvian, Russian, and Slovak). Table \ref{task_sat_data} describes volumes of data per language. The sources for this dataset are: 
\begin{itemize}
\setlength{\itemsep}{0pt}
    \item Czech: \cite{tsczh},
    \item Latvian: latvian-tweet-sentiment-corpus \footnote{https://github.com/FnTm/latvian-tweet-sentiment-corpus},
    \item Russian: mokoron\footnote{http://study.mokoron.com/},
    \item Slovak: Sentigrade\footnote{https://sentigrade.fiit.stuba.sk/data}.
\end{itemize}

As previously, we used classification accuracy as our main metric and evaluated all possible directions of transfer learning for three multilingual models (mBERT, mDBERT, and XLM-R). Check table \ref{task_sat_res} for the results, more details are available on W\&B\footnote{https://wandb.ai/eenlp/twit\_sentiment}
\\\\
\begin{table}[]
\begin{tabular}{llllll}
& & \textbf{ru}& \textbf{sk}& \textbf{lv}& \textbf{cz}\\
 & mDBERT & 88\% & 61\% & 75\% & 68\% \\
\textbf{ru} & mBERT & 89\% & 60\% & 76\% & 67\% \\
 & XLMR & \textbf{95}\% & \textbf{72}\% & \textbf{78}\% & \textbf{79}\% \\
 & mDBERT & 72\% & 51\% & 30\% & 35\% \\
\textbf{sk} & mBERT & \textbf{76}\% & 59\% & 32\% & 38\% \\
 & XLMR & 72\% & \textbf{70}\% & \textbf{38}\% & \textbf{42}\% \\
 & mDBERT & 39\% & 35\% & 70\% & 52\% \\
\textbf{lv} & mBERT & \textbf{56}\% & \textbf{38}\% & 71\% & 51\% \\
 & XLMR & 32\% & \textbf{38}\% & \textbf{73}\% & \textbf{57}\% \\
 & mDBERT & 36\% & 47\% & 60\% & 74\% \\
\textbf{cz} & mBERT & \textbf{44}\% & 50\% & 60\% & 76\% \\
 & XLMR & \textbf{44}\% & \textbf{62}\% & \textbf{71}\% & \textbf{79}\%
\end{tabular}
\caption{Twitter sentiment detection task. Validation accuracy.}
\label{task_sat_res}
\end{table}
This task results support several observations:
 \begin{itemize}
    \item again, the distilled model shows the worst ability to cross-lingual transfer learning;
    \item again, the similarity of the languages matters for the cross-lingual transfer quality;
    \item again, XLM-RoBERTa generally dominates over the rest of the models regardless of the transfer pair.
\end{itemize}   

\subsection{NLI task}

\begin{table}[h!]
\centering
\begin{tabular}{lrr}
 Language & Items & Classes   \\ \hline
 \begin{tabular}[c]{@{}l@{}}English\end{tabular}   & 30 000 & 3 \\
 \begin{tabular}[c]{@{}l@{}}Bulgarian\end{tabular}   & 30 000 & 3 \\
 \begin{tabular}[c]{@{}l@{}}Polish\end{tabular}   & 9 000 & 3 \\
 \begin{tabular}[c]{@{}l@{}}Russian\end{tabular}   & 30 000 & 3 \\
 \begin{tabular}[c]{@{}l@{}}Slovenian\end{tabular}   & 306 & 3 \\
\end{tabular}
  \caption{NLI dataset properties. Each item is a pair of sentences.}
  \label{task_nli_data}
\end{table}

Finally, we proposed the evaluation on the natural language inference (NLI) task. NLI is the task of predicting whether a hypothesis sentence is true (entailment), false (contradiction), or undetermined (neutral) given a premise sentence. Our dataset for this task also covers 4 Eastern European languages (Bulgarian, Polish, Russian, and Slovenian) + English. Table \ref{task_nli_data} describes volumes of data per language. The sources for this dataset are: 
\begin{itemize}
\setlength{\itemsep}{0pt}
    \item English: \cite{XGLUE},
    \item Bulgarian: \cite{XGLUE},
    \item Polish: \cite{KLEJ},
    \item Russian: \cite{XGLUE},
    \item Slovenian: \cite{SloveneSuperGLUE}.
\end{itemize}

As in \cite{tacl_a_00288} and \cite{nooralahzadeh-etal-2020-zero}, we used classification accuracy as the metric for the NLI task and evaluated all possible directions of transfer learning for the same three multilingual models (mBERT, mDBERT, and XLM-R). Table \ref{task_nli_res} provides the results, more details are available on W\&B\footnote{https://wandb.ai/eenlp/nli}
\\\\
\begin{table}[]
\begin{tabular}{lllllll}
& & \textbf{en}& \textbf{pl}& \textbf{ru}& \textbf{bg}& \textbf{sl}\\
            & mDBERT & 65\% & \textbf{62}\% & 56\% & 55\% & 34\% \\
\textbf{en} & mBERT  & 69\% & 48\% & 59\% & 58\% & \textbf{37}\% \\
            & XLMR   & \textbf{76}\% & 48\% & \textbf{69}\% & \textbf{72}\% & 36\% \\
 
 & mDBERT & 40\% & 92\% & \textbf{39}\% & \textbf{40}\% & \textbf{26}\% \\
\textbf{pl} & mBERT & 42\% & 92\% & 38\% & 38\% & 19\% \\
 & XLMR & \textbf{46}\% & \textbf{94}\% & 38\% & \textbf{40}\% & 14\% \\

 & mDBERT & 62\%& \textbf{57}\% & 54\% & 55\% & \textbf{33}\% \\
\textbf{ru} & mBERT & 68\% & 54\% & 54\% & 60\% & 28\% \\
 & XLMR & \textbf{75}\% & \textbf{57}\% & \textbf{64}\% & \textbf{73}\% & 31\% \\

 & mDBERT & 62\% & \textbf{62}\% & 56\% & 54\% & 31\% \\
\textbf{bg} & mBERT & 67\% & 38\% & 61\% & 52\% & \textbf{38}\% \\
 & XLMR & \textbf{74}\% & 56\% & \textbf{68}\% & \textbf{64}\% & 27\% \\


\end{tabular}
\caption{NLI task. Validation accuracy. We omitted training on Slovenian since that dataset consists only of 306 items.}
\label{task_nli_res}
\end{table}
This task results support several observations:
 \begin{itemize}
    \item this task seems to be significantly harder than others;
    \item there is no consistent winner among the models for this task.
\end{itemize}   

\section{Conclusions and discussion}

In this project, we have made and published a broad index of NLP resources for Eastern European languages, which, we hope, will be helpful for the NLP community. We have invested in the creation of new cross-lingual datasets focused on Eastern European languages, hand-crafted the benchmarks for 5 common reasoning tasks, and provided the evaluations of several modern multilingual models. We have published all our code to support future research.
\\\\
As we have highlighted in our analysis, the quality of cross-lingual transfer learning depends on various factors, including the pre-training and distillation details of the model, the similarity between the languages, the size of the dataset for the downstream task, etc. These observations are consistent with the conclusions of previous studies (check \cite{doddapaneni2021primer} for more details). The general observations from these experiments are:
\begin{itemize}
    \item the multilingual pre-training is the key factor for successful cross-lingual transfer learning;
    \item the distillation significantly decreases the quality of cross-lingual transfer learning;
    \item the cross-lingual transfer quality seems to be correlated with the degree of language affinity;
    \item XLM-RoBERTa generally dominates over the rest of the tested models.
\end{itemize}   

We are considering two major directions for future work in the scope of better understanding cross-lingual abilities of the modern models in the context of less-resourced Eastern European languages:
\begin{itemize}
    \item leveraging the synthetic datasets and data augmentation methods for supporting the less-resourced languages;
    \item and evaluation of the less general models (for example, models like CroSloEngual\cite{CroSlo}, BERTić \cite{BERTIC}, SlavicBERT \cite{SlavicBERT}, or LitLat BERT).
\end{itemize}   
\section{Acknowledgements}
This work was initially started as an EEML 2021 summer school project. We would like to thank all the organizers, especially Viorica Patraucean and Razvan Pascanu for their valuable support. We also would like to thank our mentor Tomáš Mikolov for wise advice and recommendations.

\section{Bibliographical References}\label{reference}

\bibliographystyle{lrec2022-bib}
\bibliography{lrec2022-example}

\bibliographystylelanguageresource{lrec2022-bib}

\end{document}